\documentclass[11pt]{article}

\usepackage{enumitem}

\usepackage[final]{acl}

\usepackage{times}
\usepackage{latexsym}

\usepackage[T1]{fontenc}
\usepackage[utf8]{inputenc}

\usepackage{microtype}

\usepackage{inconsolata}

\usepackage{graphicx}

\usepackage{multirow}%
\usepackage{amsmath,amssymb,amsfonts}%
\usepackage{amsthm}%
\usepackage{mathrsfs}%
\usepackage[title]{appendix}%
\usepackage{xcolor}%
\usepackage{textcomp}%
\usepackage{manyfoot}%
\usepackage{booktabs}%
\usepackage[linesnumbered,ruled,vlined]{algorithm2e}
\usepackage{listings}%
\usepackage{afterpage}
\usepackage{todonotes}
\usepackage{enumitem}

\usepackage[most]{tcolorbox}
\usepackage{tikz}
\usetikzlibrary{positioning}
\usepackage{graphicx}
\usepackage{amsmath}
\usepackage{float}
\usepackage{booktabs}
\usepackage{graphicx}
\usepackage[colorlinks=true, allcolors=blue]{hyperref}
\usepackage{makecell}
\usepackage{fancyvrb}
\usepackage{multicol}
\usepackage[ruled]{algorithm2e}
\usepackage{amsmath,amssymb}
\usepackage{spverbatim}
\usepackage{tcolorbox}
\tcbuselibrary{skins,breakable}
\tcbset{enhanced jigsaw, breakable}
\usepackage{multirow}
\usepackage{caption} 
\usepackage{subcaption}
\title{A Multi-Stage Agentic Framework for Effective Counter-Narrative Generation and Refinement}

\author{
 \textbf{Carmel Kronfeld\textsuperscript{1}},
 \textbf{Sharva Gogawale\textsuperscript{2}},
 \textbf{Tetsuro Kobayashi\textsuperscript{3}},
 \textbf{Irad Ben-Gal\textsuperscript{1}}
\\
\\
 \textsuperscript{1}School of Industrial \& Intelligent Systems Engineering, Tel Aviv University \\
 \textsuperscript{2}School of Electrical and Computer Engineering, Tel Aviv University \\
 \textsuperscript{3}School of Political Science and Economics, Waseda University
\\
 \small{
 \texttt{carmelk@mail.tau.ac.il},
 \texttt{sharvag@mail.tau.ac.il},
 \texttt{tkobayas@waseda.jp},
 \texttt{bengal@tau.ac.il}
 }
}

\begin{document}
\maketitle
\begin{abstract}
The rapid diffusion of hate speech and misinformation on social networks challenges democratic societies, since direct suppression efforts may deepen polarization, fuel public distrusts, and strengthen extremist narratives. LLM-driven counter-narratives (CNs) offer a promising way to reduce those risks, yet their effectiveness depends on rhetorical and stylistic choices that remain poorly understood. We present a multi-stage agent-based framework for generating, refining, and evaluating CNs, applied to pro-Russian hate and misinformation narratives on the war with Ukraine and adaptable to other domains. A pilot experiment with human evaluators identifies effective technique style pairings, such as \emph{repetition} with \emph{emotional} framing enhancing \emph{persuasiveness}. Building on these insights, we introduce a multi-agent refinement process that iteratively improves CNs for \textit{persuasiveness}, \textit{emotional engagement}, and \textit{shareability}. After human validation confirmed improvement, an automated safety analysis shows that our refined CNs match or improve on expert-written counterspeech. A simulated experiment then shows that they reduce the perceived strength of pro-Russian narratives and consistently outperform a vanilla LLM baseline, highlighting a pathway toward scalable, narrative-specific interventions against hate speech and misinformation. Code and data accompanying this work are publicly available at \url{https://github.com/carmelkron/inlg2026-counter-narratives}.
\end{abstract}

\section{Introduction}

With the global diffusion of social media, cross-border propaganda has gained unprecedented influence. Unlike earlier forms that relied on state media or diplomatic channels, today’s influence campaigns follow a “participatory propaganda” model, where narratives spread through state-linked propagators, astroturfing accounts, bots, and even ordinary users acting as secondary disseminators \citep{Starbird2019, Wanless2021}.

A key driver of this shift is the growth of coordinated mis/disinformation operations that exploit platform dynamics at scale. By synchronizing amplification signals (e.g., likes, shares, and comments) via bot networks and inauthentic coordination, they can game ranking algorithms, manufacture the appearance of consensus, and manipulate public discourse; reflecting this severity, the World Economic Forum ranks misinformation and disinformation as the most severe near-term global risk, citing harms such as polarization and eroding trust in institutions \citep{WEF2024, WEF2025}.
These campaigns succeed not only by spreading falsehoods, but by optimizing for attention: emotionally charged messages often outweigh fact-based ones \citep{Xue2025}, and many claims remain persuasive even when unfounded or conspiratorial \citep{Kobayashi2025}. In parallel, hate speech frequently co-propagates with misinformation, normalizing hostility toward out-groups and, in some settings, showing measurable links to offline hate incidents and violence \citep{CASTANOPULGARIN2021101608, arcila2024online}.

Recent work further suggests that the persuasive edge of online narratives may increase as generative AI becomes more capable. \citet{hackenburg2025leverspoliticalpersuasionconversational} show that LLMs can enhance persuasiveness through strategic selection and framing of information, but this may reduce factual accuracy, making arguments more convincing but less truthful. Moreover, language barriers have decreased with AI, enabling fluent localized messaging and expanding narrative reach \citep{Wack2025}.

Illiberal narratives play a particularly significant role in international conflicts, where influence over global public opinion can shape institutional resolutions, negotiations, and coalition building. Accordingly, conflict parties conduct wide-ranging social media campaigns, including in the Russia-Ukraine context. Empirical studies demonstrate that bots and automated accounts played a crucial role in the early stages of diffusion of these narratives \citep{Geissler2022}.

\begin{figure*}[h]
    \centering
\includegraphics[width=0.7\textwidth]{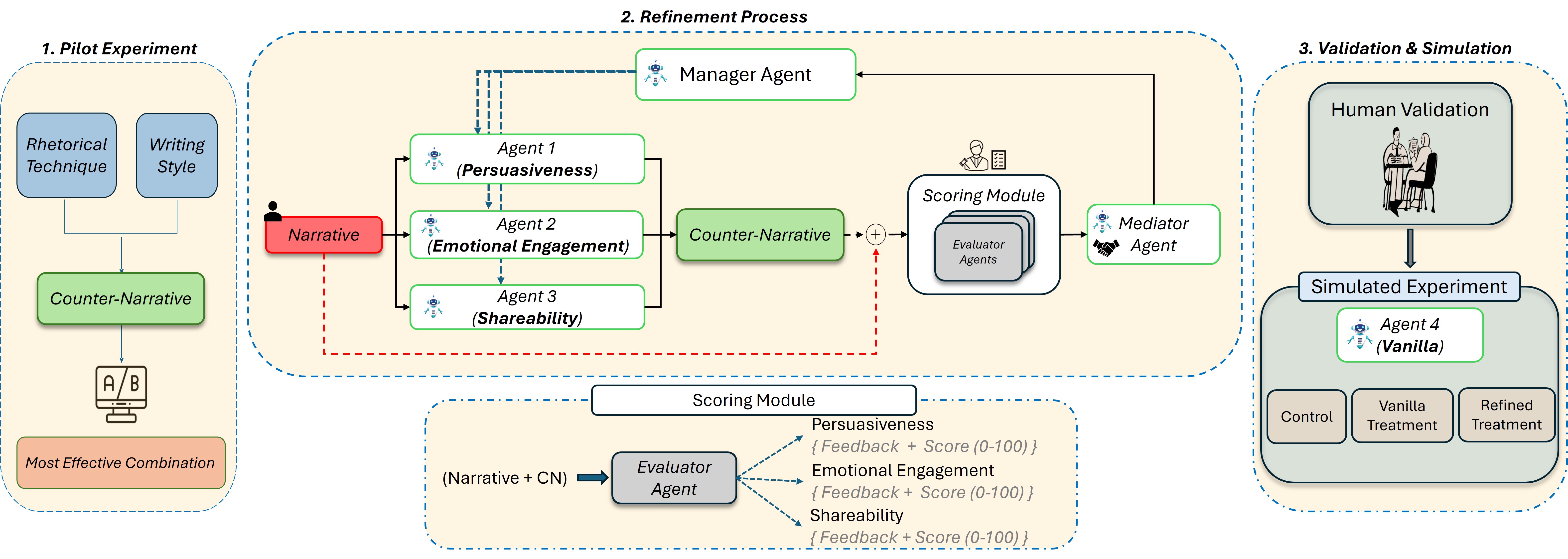}
    \caption{The proposed multi-stage framework.}
    \label{fig:workflow}
\end{figure*}

Taken together, the rapid spread of misinformation and hateful, emotionally charged narratives poses major challenges, yet in democratic societies direct state intervention is difficult: online censorship can undermine freedom of expression, and suppressing speech tied to particular political positions may deepen polarization, fuel distrust, and strengthen extremist narratives. Accordingly, counter-narratives (CNs) have gained attention as a way to protect a democratic and liberal public sphere without relying on direct interventions such as censorship. CNs go beyond fact-checking by re-framing narratives to reduce their persuasive and emotional impact \citep{Cipers2023}. 
More broadly, recent defense work frames such campaigns as “cognitive warfare” and emphasizes mitigation and societal resilience, rather than relying solely on takedowns, motivating scalable “more-speech” interventions like counter-narratives \citep{blatny_sondergaard_2025_cognitive_warfare}.
However, manual CN creation is slow and effort-intensive, limiting scalability \citep{schieb2016governing}, motivating automated LLM-based CN generation, although little is known about which stylistic and rhetorical strategies are most effective.
We address these gaps by proposing a multi-stage agent-based framework that unifies CN generation, refinement, and evaluation, enabling scalable, narrative-specific interventions. Our contributions are as follows:
\begin{itemize}[leftmargin=*, noitemsep, topsep=0pt]

\item A hybrid human-AI pipeline integrating automated CN generation, refinement, and evaluation as shown in Figure~\ref{fig:workflow}. 

\item A pilot study with human evaluators studying which \emph{rhetorical techniques} and \emph{writing styles} are most effective for CN generation.

\item A multi-agent refinement system aiming to improve CN generation by effectively leveraging leading technique-style pairs, showing clear improvements, validated by human evaluators and shown to be comparable to expert-written counterspeech on the automated harmful-language indicators evaluated.

\item A simulated experiment showing that our refined CNs outperform a vanilla LLM baseline in reducing the effectiveness of the target narrative.

\end{itemize}

% \item Empirical validation through a pilot study, large-scale refinement of 20 narratives and human evaluation. In particular, studying how \emph{rhetorical techniques} and \emph{writing styles} can be effectively leveraged in CNs. We also conducted a simulated experiment with generative agents showing that our refined CNs outperform vanilla baselines.

Although our case study centers on harmful and misleading pro-Russian narratives, the framework generalizes to other domains, underscoring its broader potential for scalable LLM-driven CNs development.

\section{Related Work}

\paragraph{CN Generation.} The strategy of “more speech” to counter hate speech \citep{bielefeldt2011ohchr} underlies both counter-speech and CNs, often treated interchangeably in NLP, though distinguished in social sciences \citep{ruths2016counterspeech}. Early work centered on data curation, building datasets from social media \citep{mathew2019thou} and expert replies \citep{chung2019conan}, later expanded with multi-target, human-in-the-loop pipelines \citep{fanton2021human}. These enabled early generative approaches like generate-prune-select \citep{zhu2021generate}.

Evidence that counterspeech can shift norms \citep{schieb2016governing} spurred use of LLMs to create more persuasive, evidence-grounded CNs. Factuality has been addressed through reinforcement learning rewards (F2RL, \citealt{wang2024f2rl}), retrieval augmentation (ReZG, \citealt{jiang2023raucg}), and attention regularization \citep{bonaldi2023weigh}. Strategic control has leveraged intent conditioning (e.g., empathy, denouncement), with dual discriminators in DART \citep{wang2024dart}, and preference-based tuning such as DPO for improved factuality and multilingual robustness \citep{wadhwa2024northeastern}. Argumentative information \citep{furman2023high} and personalization \citep{dougancc2023generic} have also enhanced CN quality.

Analyzing the linguistic characteristics of untrustworthy text, \citet{rashkin} demonstrated that stylistic cues systematically distinguish propaganda and hoaxes from reliable news. In the realm of computational argumentation generation, \citet{alshomary2021} introduced multi-stage pipelines that first identify weak premises before generating targeted counter-arguments, while \citet{alshomary2022} explored style-conditioned persuasion using moral framing.

Recent research explores prompting strategies \citep{jeong2025large, saha2024zero}, emotional framing \citep{russo2023countering}, and real-world deployments, e.g., countering hate toward Ukrainian refugees \citep{podolak2023analyzing}. Risks remain, as \citet{bar2024generative} caution LLM counterspeech may backfire by amplifying harmful narratives. Applications now extend beyond hate speech to misinformation, using control codes in argument-graph frameworks \citep{saha_mis} and scalable fact-grounded generation \citep{xu2025generating}. Overall, the field has moved from curated-data pipelines to LLMs producing safer, context-specific CNs.

\paragraph{Persuasion using LLMs.} Research shows that newer LLMs exhibit strong persuasive capabilities, producing arguments comparable to human-written ones \citep{durmus2024persuasion}. They scale personalized persuasion \citep{matz2024potential}, outperform humans in controlled debates with sociodemographic access \citep{salvi2025conversational}, and shift opinions even in naturalistic conversations where users know they are interacting with AI \citep{havin2025can}. Persuasiveness is linked not only to argument quality but also to replicating nuanced communicative intent \citep{donmez2025understand}.

To advance these systems, \citet{jin2024persuading} introduced DailyPersuasion and PersuGPT, combining intent-to-strategy reasoning with optimization, while \citet{ramani2024persuasion} proposed a multi-agent architecture with auxiliary agents for strategy and fact-checking. These methods enhance efficacy and show applications in propaganda \citep{goldstein2024persuasive} and influencing polarized political attitudes \citep{bai2025ai}. However, these new methods raise significant risks. For example, \citet{liu2025llm} warn that LLMs can be dangerous persuaders, while \citet{bozdag2025persuade} emphasize systematic evaluation of persuasion effectiveness and susceptibility. Together, these works highlight both the promise and the dangers of persuasive LLMs.

\paragraph{CN Evaluation.} Evaluating CNs is challenging, as traditional reference-based metrics correlate poorly with human judgments and overlook qualities like coherence and contextual relevance. This has led to a shift toward LLM-based, reference-free evaluation.
One paradigm decomposes quality into human-centric dimensions. \citet{jones2024multi} rate CNs on five NGO-inspired aspects, aligning with human annotations; CSEval \citep{hengle2025cseval} applies auto-calibrated Chain-of-Thought to four dimensions including aggressiveness and suitability; \citet{song2024assessing} add “human-likeness” for naturalness; and CheckEval \citep{lee-etal-2025-checkeval} stresses evaluators via checklist-driven tests.
A second paradigm relies on comparative ranking. \citet{zubiaga2024llm} introduce a tournament pipeline where LLM judges rank CNs pairwise, achieving strong human correlation. Extensions evaluate persuasion and factuality, yielding richer feedback for generation refinement \citep{wilk2025fact}.

\section{Pilot Experiment}

Our pilot study maps which \emph{rhetorical techniques} and \emph{writing styles} are most effective for constructing CNs against prominent pro-Russian narratives. We (i) extract representative base claims from pro-Russian Twitter/X discourse; (ii) systematically generate CNs by crossing 13 rhetorical techniques with 10 styles; and (iii) obtain human judgments on three key performance indicators (KPIs): \emph{persuasiveness}, \emph{emotional engagement}, and \emph{shareability}. The technique taxonomy follows previous research \citep{chernyavskiy2024zenpropaganda, dasanmartino2019finegrained, salman2023detecting}. Prompts used for generation are provided verbatim in Appendix~\ref{app:prompts}, and the evaluation interface is shown in Appendix~\ref{app:interface}.

\subsection{Extraction of Pro-Russian Base Claims}
On April 2024, in collaboration with XPOZ,\footnote{\url{https://www.xpoz.ai/}} we collected English-language tweets containing keywords that reflect common misinformation tropes and hate-inciting framings targeting Ukraine (e.g., ``Ukraine war crimes'' and ``Ukraine Nazi'') spanning up to 120 days prior to retrieval. After filtering noise (e.g., off-topic content), we clustered \(\sim\)10{,}000 tweets with HDBSCAN and consolidated the content into three high-frequency base claims in tweet-like form (see Appendix~\ref{base:claims}). These three claims capture recurring narratives and serve as inputs for CN generation.

\subsection{Construction of Counter-Narratives}
We hypothesize that both technique and style shape CN effectiveness. Together they operationalize how the CN is expressed, shaping its framing, emphasis, tone, and argumentative structure. We therefore cross 13 techniques with 10 styles: \textit{Pessimistic, Optimistic, Emotional, Rational, Dry language, Metaphorical, Amusing, Cynical, Empathic, Detached}. Using \emph{LLAMA-3.1-70B-Versatile} as an LLM, for each of the three base claims, we generated one CN using each technique-style pair, yielding \(3 \times 13 \times 10 = 390\) CNs.

\subsection{Human Evaluation of Counter-Narratives}
We evaluated the \(390\) CNs on three KPIs: (1) \emph{persuasiveness}, the primary criterion for counterspeech; (2) \emph{emotional engagement}, motivated by evidence that emotionally engaging social media content receives higher interaction than factual messaging \citep{Xue2025}; and (3) \emph{shareability} (likelihood to repost/retweet), reflecting the importance of diffusion dynamics in online settings.

Five native English-speaking undergraduate political science majors from Waseda University served as judges. Political science majors were selected to ensure sufficient background knowledge of the Russia-Ukraine context for interpreting claims and CNs. The evaluation interface presented the base claim and two CNs (A and B) side-by-side. For each KPI, annotators made forced-choice pairwise comparisons: “Which is more convincing (A/B)?”, “Which evokes stronger emotions (A/B)?”, “Which is more likely to be shared (A/B)?”. We adopt pairwise comparisons to mitigate scale-use bias and to sharpen relative signal among closely ranked CNs.

\subsection{Identification of Effective Combinations}
To identify which technique-style combinations drive outcomes, we model the pairwise choices using LASSO-regularized logistic regression. The model includes the techniques, styles, and their interactions for the two CNs being compared. After regularization, we rank the selected technique-style interactions by coefficient magnitude. Table~\ref{tab:best_combinations} reports the top combination for each KPI.  Evaluator internal consistency and inter-rater agreement are reported in Appendix~\ref{app:agreement}.

\begin{table}[h!]
\small
\centering
\renewcommand{\arraystretch}{1.15}
\begin{tabular}{p{0.27\linewidth} p{0.35\linewidth} p{0.23\linewidth}}
\hline
\textbf{KPI} &
\textbf{\begin{tabular}[c]{@{}l@{}}Rhetorical\\technique\end{tabular}} &
\textbf{Writing style} \\
\hline
Persuasiveness & Repetition & Emotional \\
\hline
\begin{tabular}[l]{@{}l@{}}Engagement\\Emotional\end{tabular} & Fear mongering & Empathic \\
\hline
Shareability & Card stacking & Metaphorical \\
\hline
\end{tabular}
\caption{Best combination of rhetorical technique and writing style per KPI}
\label{tab:best_combinations}
\end{table}

\section{Refinement Per Claim Process}
\label{refinement}

The pilot experiment identified effective technique-style combinations for CNs, but relied on simple prompts and a small set of claims, limiting nuance and generalizability. Building on these findings, the refinement process adds a second layer of specialization: beyond using the best technique-style combinations from the pilot, it iteratively refines narrative-specific CN Generator prompts for each narrative and target KPI, allowing the agents to adapt their rhetorical and stylistic behavior to the unique content and framing of each narrative.
The process extends to twenty representative hate and misinformation pro-Russian narratives drawn from a large number of tweets (see Appendix~\ref{app:twenty_claims}), producing generators optimized for \emph{persuasiveness}, \emph{emotional engagement}, or \emph{shareability}. For each narrative, three specialized generators are developed, yielding 60 agents that can be flexibly deployed depending on the application context. Because large-scale human evaluation is infeasible, the process relies on LLM-as-judge strategies, where evaluator agents impersonate pro-Russian users, enabling iterative prompt refinement without prohibitive cost. Such an approach resembles Generative Adversarial Networks (GANs) as a way to train a generator to produce new data (image, audio, text, etc.) by pitting it against a discriminator in a two-player game \citep{10.1145/3422622}. 
We deliberately chose pro-Russian evaluator agents as a hard adversary to stress-test CNs under a hostile audience model. The intuition is that if a CN is rated persuasive, emotionally engaging, or shareable even by strongly pro-Russian accounts, after refinement forces it to address the strongest objections and narrative defenses, it is more likely to remain competitive for less-committed, peripheral audiences.

\subsection{Architecture Overview}
The system is implemented as a multi-agent architecture built on the \textit{SmolAgents} \citep{smolagents} framework with \textit{Claude-3.5-Haiku}\footnote{\href{https://assets.anthropic.com/m/1cd9d098ac3e6467/original/Claude-3-Model-Card-October-Addendum.pdf}{Model Card Addendum: Claude 3.5 Haiku and Upgraded Claude 3.5 Sonnet}} as the backbone LLM. It distributes responsibilities across five components. 
\textbf{3 CN Generator Agents} produce narrative-specific outputs using assigned rhetorical techniques and writing styles, each tied to one KPI. Each agent is initialized with the best-performing technique-style combination identified for its corresponding KPI in the pilot experiment, ensuring that generation begins from the most effective rhetorical foundation.
\textbf{18 Pro-Russian Evaluator Agents} impersonating 18 authentic pro-Russian X users chosen in two steps: (i) identifying accounts repeatedly posting pro-Russian narratives in our tweet dataset, then (ii) filtering these accounts by high engagement (views) and high likelihood of a genuinely pro-Russian stance. Each persona is grounded with three layers: a detailed LLM summary of the user's tweet history, ten positive tweet examples from that user to anchor voice, and ten negative tweet examples from other pro-Russian users to sharpen distinctiveness. Crucially, the fidelity of this impersonation is empirically validated in Appendix~\ref{app:impersonation}. On every narrative-CN pair, evaluators return KPI scores on a 0-100 scale and structured feedback listing both good and bad points for each KPI. Evaluators maintain persistent memory across iterations, so feedback and scoring evolve consistently with the persona’s prior judgments.
A \textbf{Mediator Agent} consolidates evaluator comments into exactly five representative good points and five representative bad points per KPI, eliminating redundancy and balancing divergent views.
A \textbf{Manager Agent} then revises generator prompts, integrating aggregated feedback and KPI statistics while preserving each agent's rhetorical configuration; it maintains a rolling memory of past refinements to avoid repeating ineffective adjustments.
Finally, a \textbf{Memory Summarizer Agent} compresses evaluator histories into concise summaries, ensuring continuity without exceeding context limits. Together, these components emulate an audience-response loop that enables systematic refinement at scale.
All prompts used in this step appear in Appendix~\ref{app:rpc_prompts}.

\subsection{The Process}
For each pro-Russian narrative, three CN Generator Agents are initialized alongside the Mediator, Manager, and Memory Summarizer. The process proceeds sequentially, refining one generator at a time. At each iteration, the active Generator produces a CN, all Evaluator Agents score it and provide feedback, and every five iterations the Memory Summarizer compresses histories to preserve long-term context. The Mediator aggregates these outputs into representative good and bad points per KPI, while score statistics (means and variances) are computed in parallel. The Manager integrates this information with the generator’s prompt, updating it to better target the designated KPI while retaining the assigned technique-style pair. In our runs, we stop at 25 iterations or early-stop after 8 rounds without improvement, then repeat for the next KPI until all three agents are optimized. Pseudo-code is in Algorithm~\ref{alg:rpc_single_claim}.

\subsection{Results Analysis}

Our analysis starts from the observation that pro-Russian narratives recur as variations of a few meta-narratives. Grouping the 20 narratives into meta-narratives reduces noise and mirrors how narratives cluster on social media, allowing trend analysis at the thematic level. This structure allows us to test whether some themes are systematically more or less susceptible to CNs and identify where differences are most pronounced across our KPIs. Specifically, we identified six meta-narratives, with the number of narratives assigned to each noted in parentheses: 

\begin{enumerate}[leftmargin=*, noitemsep, topsep=0pt]
    \item NATO/Western Aggression \& Broken Promises (5)
    \item Western Manipulation, Hegemony \& Moral Decay (5)
    \item Humanitarian/ ''Denazification'' Justifications \& Ukraine’s Wrongdoing (5)
    \item Military Success \& Liberation Narrative (2)
    \item Territorial Legitimacy via Referendum (1)
    \item Economic Warfare \& War-Profit Claims (2)
\end{enumerate}

Using these groups, we then examine refinement dynamics to determine which families are harder to counter and along which dimensions performance diverges. We now turn to the specific analyses conducted, each accompanied by an illustrative example, before concluding with a discussion of the broader insights drawn across all analyses.

\begin{figure*}[h]
    \centering
    \begin{subfigure}{0.3\textwidth}
        \includegraphics[width=\linewidth]{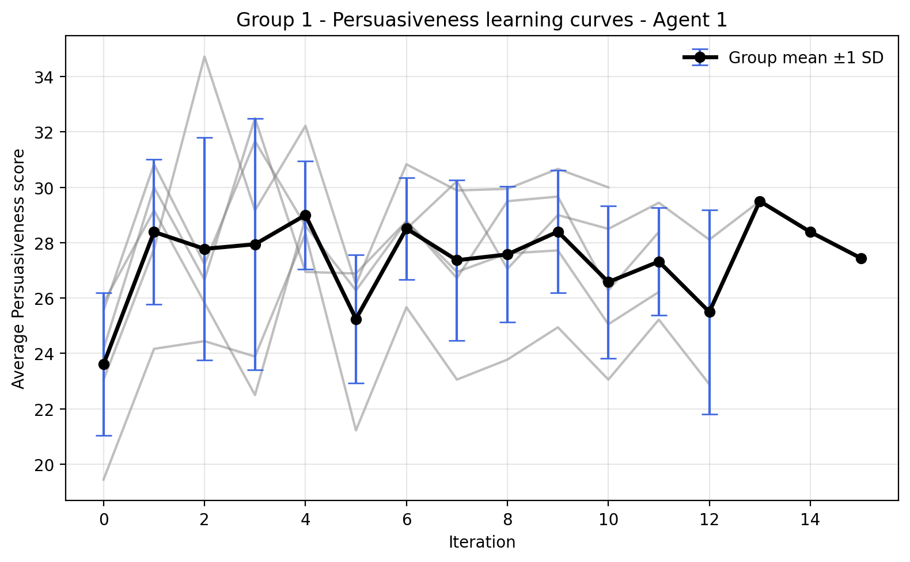}
        \caption{Persuasiveness}
    \end{subfigure}
    \hfill
    \begin{subfigure}{0.3\textwidth}
        \includegraphics[width=\linewidth]{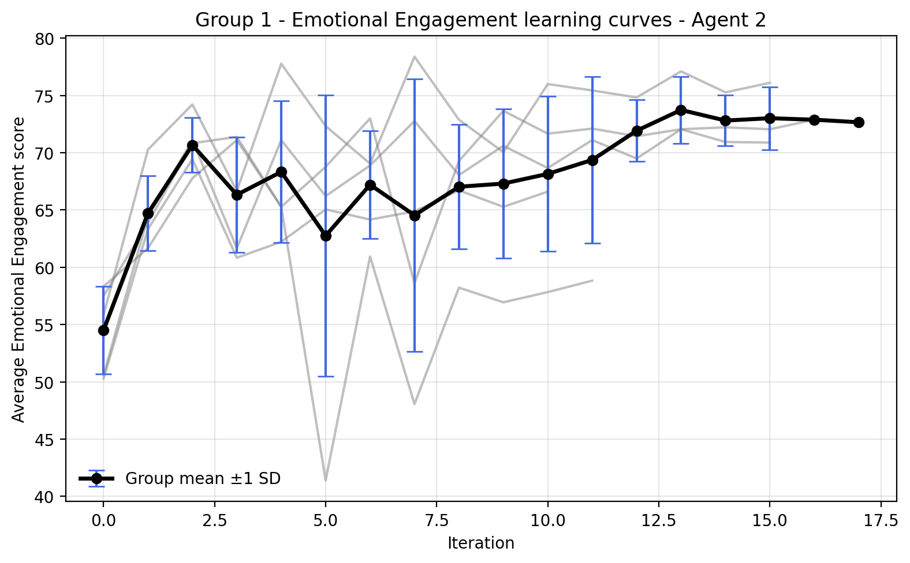}
        \caption{Emotional Engagement}
    \end{subfigure}
    \hfill
    \begin{subfigure}{0.3\textwidth}
        \includegraphics[width=\linewidth]{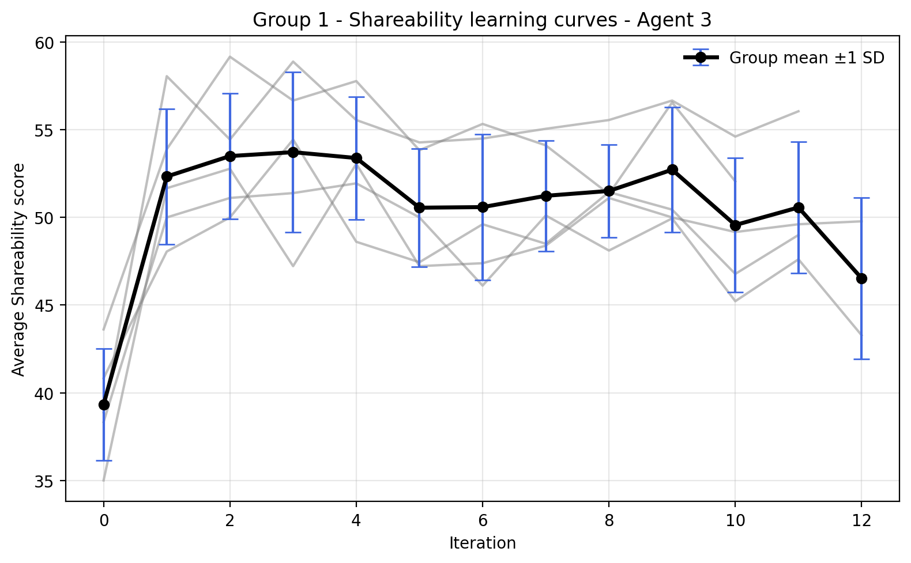}
        \caption{Shareability}
    \end{subfigure}
    \caption{Refinement curves for Group 1 (NATO / Western Aggression \& Broken Promises) across the three KPIs. Black lines represent group means; blue bars indicate $\pm 1$ standard deviation.}
    \label{fig:group1_curves}
\end{figure*}

\paragraph{Refinement Curves per Group:}  
For each meta-narrative, we plotted curves of average KPI scores across iterations to assess learning dynamics such as stability, rate of improvement, and plateauing. Figure~\ref{fig:group1_curves} illustrates Group~1. \textit{persuasiveness} appears volatile, fluctuating without steady gains; \textit{emotional engagement} rises more consistently before stabilizing; and \textit{shareability} increases sharply early on, then plateaus and slightly declines. These trajectories highlight how refinement outcomes might differ by KPI within a narrative group.

\begin{figure}[h!]
    \centering
    \includegraphics[width=0.7\linewidth]{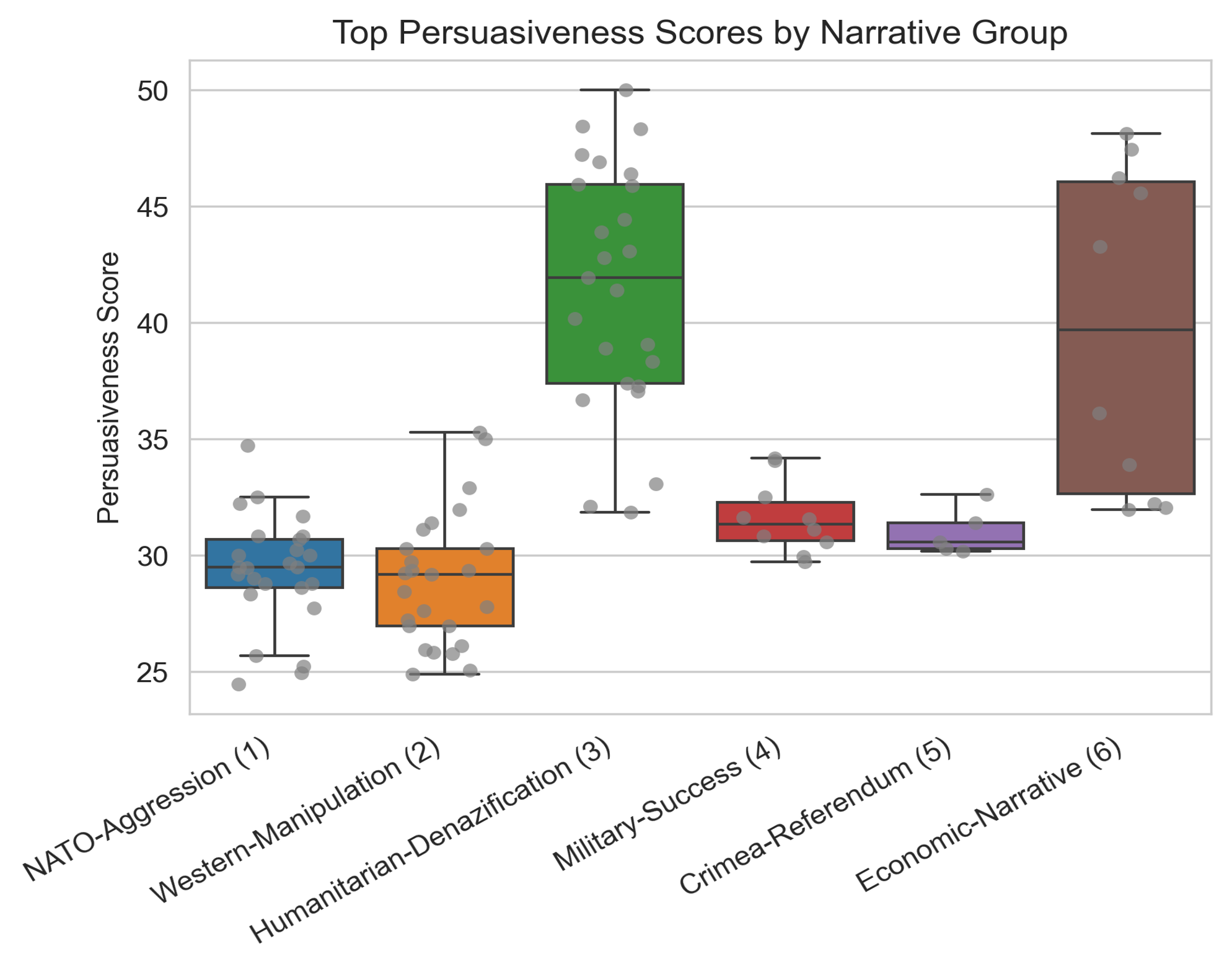}
    \caption{Peak persuasiveness scores achieved, grouped by meta-narrative. Each box represents the distribution of peak values across all narratives in that group.}
    \label{fig:peak_scores_persuasion}
\end{figure}

\paragraph{Peak Achievable Scores:}  
To estimate the maximum effectiveness of the refinements, we measured each group’s peak performance of each KPI, measured by taking the top five scores per narrative to balance robustness against outliers and avoid diluting maxima. We compared groups using a Kruskal-Wallis test followed by Dunn tests with Holm correction. Figure~\ref{fig:peak_scores_persuasion} shows peak persuasiveness distributions, and Table~\ref{tab:kw_dunn} reports test results. Groups~3 and 6 reached significantly higher peaks than Groups~1 and 2, with Group~3 also outperforming Group~4. By contrast, Groups~1 and 2 clustered at consistently low levels, while Groups~4 and 5 remained in intermediate ranges. These findings indicate a systematic variation in achievable \textit{persuasiveness} across narrative families.

\begin{table}[h!]
\small
\centering
\begin{tabular}{p{0.5\linewidth} p{0.35\linewidth}}
\toprule
\textbf{Comparison} & \textbf{\begin{tabular}[c]{@{}c@{}}p-value\\(Holm-corrected)\end{tabular}} \\
\midrule
Kruskal-Wallis (overall) & $p = 5.47 \times 10^{-13}$ \\
\midrule
3 vs. 2 & $p = 3.93 \times 10^{-10}$ \\
3 vs. 1 & $p = 2.66 \times 10^{-9}$ \\
6 vs. 2 & $p = 6.82 \times 10^{-5}$ \\
6 vs. 1 & $p = 1.79 \times 10^{-4}$ \\
3 vs. 4 & $p = 0.036$ \\
\bottomrule
\end{tabular}
\caption{Statistical testing of peak persuasiveness scores across meta-narrative groups.}
\label{tab:kw_dunn}
\end{table}

\begin{figure}[h!]
    \centering
    \includegraphics[width=1\linewidth]{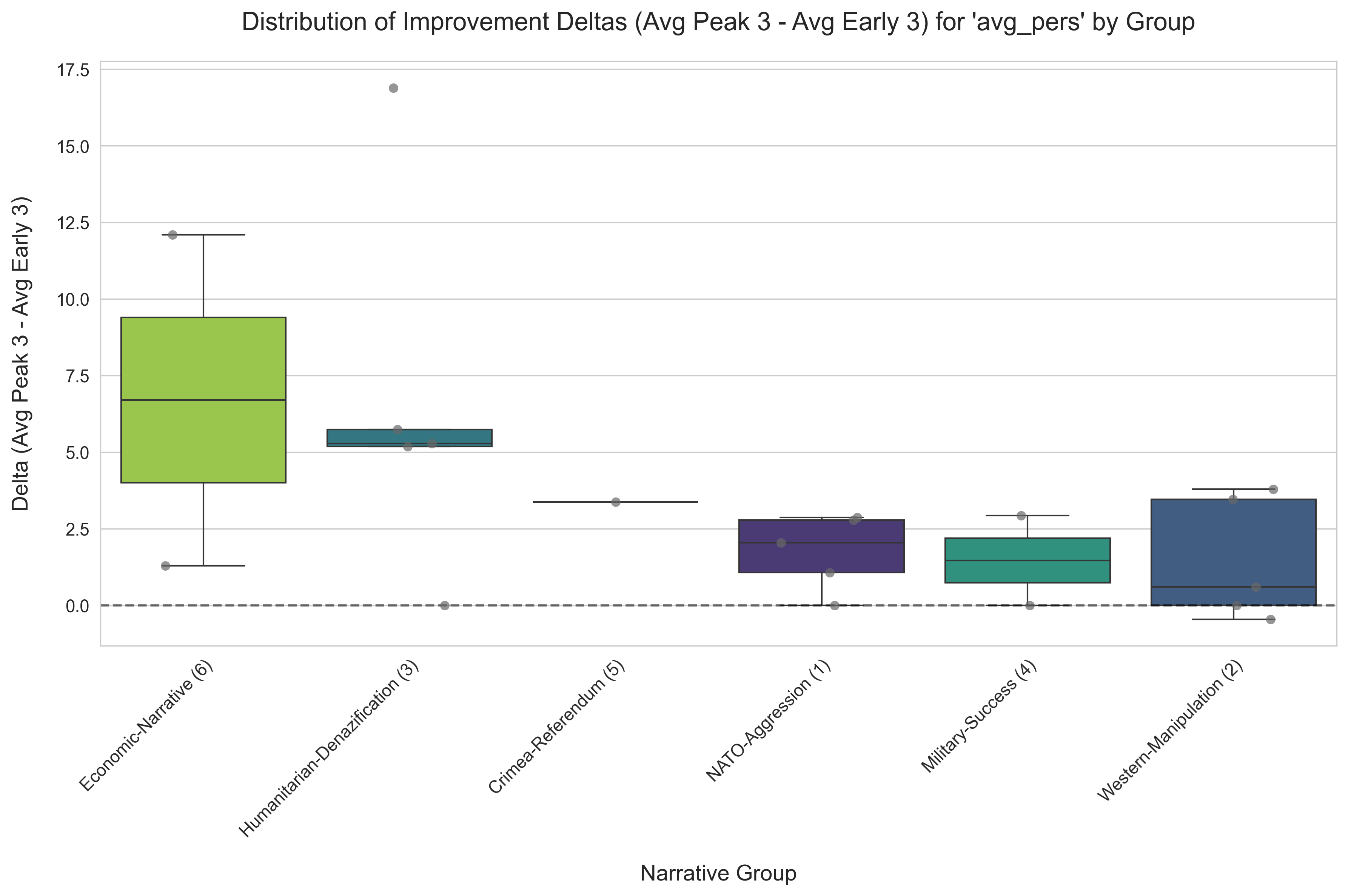}
    \caption{Improvement deltas for persuasiveness across narrative groups. Each box represents the distribution of deltas across claims in that group.}
    \label{fig:improvement_deltas_persuasion}
\end{figure}

\paragraph{Improvement Deltas:}  
To assess refinement efficiency, we compared early performance (first three iterations) with peak performance (three-iteration window around the maximum). Figure~\ref{fig:improvement_deltas_persuasion} shows the results for \textit{persuasiveness}. Group~6 achieved the largest gains, followed by Group~3, while Groups~1, 2, 4, and 5 showed modest improvements near zero. A Kruskal-Wallis test confirmed that these differences were not statistically significant ($H=5.72$, $p=0.334$), likely due to the small number of narratives per group. Thus, while descriptive patterns suggest that some families benefit more from refinement than others, these differences cannot be reliably distinguished statistically.

\subsection{Key Insights}

First, \textbf{narrative theme strongly influences achievable performance levels} for \emph{persuasiveness} and \emph{shareability}, while \emph{emotional engagement} emerges as a higher, more universal quality, indicating that affective resonance is less constrained by content and more dependent on stylistic refinement. Certain themes consistently produce CNs that are both more persuasive and shareable, while \emph{emotional engagement} reaches similarly high levels across all themes.
Second, the analysis identifies \textbf{clear leaders among narrative groups}. CNs addressing narratives in Groups 3 and 6 consistently achieve higher peak scores, indicating that Humanitarian and Economic themes are especially fertile ground for effective CNs.
Third, results demonstrate \textbf{theme-dependent score ceilings despite consistent learning efficiency}. Gains accumulate at similar rates across groups, yet ultimate peaks differ, suggesting that the process is comparably effective but bounded by thematic affordances.  
Finally, the analysis shows that \textbf{\emph{persuasiveness} remains the most volatile and difficult KPI to optimize}. Learning curves are unstable and plateau at lower ceilings than \emph{shareability} or \emph{emotional engagement}, underscoring the difficulty of consistently crafting persuasive CNs, which is natural given that the evaluator agents accurately impersonate pro-Russian users who are likely resistant to attitude change.

\subsection{Human Validation of Refinement}

Since the refinement loop is driven entirely by agent-assigned KPI scores produced by an LLM-based judge, it is necessary to verify that these scores correspond to human preferences. We conducted an initial validation study in which the two most consistent evaluators from pilot experiment judged 360 CN pairs constructed from within the refinement process, using the higher agent score as the gold label. Pairs were defined by the magnitude of the agent-assigned score gap: \emph{high-difference} pairs contained two CNs with a large score gap, and \emph{low-difference} pairs contained two CNs with a small score gap, yielding three pairs of each type per KPI per narrative. Alignment was observed only for high-difference \emph{persuasiveness} pairs, while \emph{emotional engagement} and \emph{shareability} showed weak or no alignment. Qualitative feedback revealed that many CNs contained unnatural phrasing, grammatical errors, incoherent metaphors, and misused emojis, which obscured the quality signal for these two KPIs. This initial study served as a diagnostic step: its findings directly informed the addition of naturalness and coherence guidelines to the refined prompts, targeting grammatical correctness, logical plausibility, consistent internal imagery, and authentic social-media register. Examples of CNs before and after this change can be seen in Appendices~\ref{app:cn_examples_1},\ref{app:cn_examples_2}.

Following these prompt revisions, we conducted a formal human validation designed to test whether the quality difference between refined and non-refined CNs is perceptible to human judges. Crucially, this validation redefines what \emph{high-} and \emph{low-difference} mean relative to the initial study. Rather than being determined by the magnitude of score gaps, pair type now reflects the source of the CNs: \emph{high-difference} pairs consist of one CN generated by the refined prompt and one by the non-refined prompt, making the quality gap structural and systematic; \emph{low-difference} pairs consist of either two refined or two non-refined CNs, where both options are of comparable quality by construction. For each of the 20 narratives and each of the three KPIs, we constructed four pairs of each type, yielding 480 pairs in total, and is motivated by the following logic: if the refinement process produces meaningfully better CNs, evaluators should consistently prefer the refined CN in high-difference pairs, yet show no systematic preference in low-difference pairs where both CNs are of comparable quality. The same two evaluators each judged all 480 pairs, yielding 960 judgments in total. Results confirm both predictions. Table~\ref{tab:human_val_high} reports high-difference results across KPIs.

\begin{table}[h]
\centering\small
\begin{tabular}{lcc}
\toprule
KPI & Accuracy & Cohen's $\kappa$ \\
\midrule
Persuasiveness       & 81.87\% & 0.638 \\
Emotional Engagement & 80.63\% & 0.606 \\
Shareability         & 83.75\% & 0.675 \\
\bottomrule
\end{tabular}
\caption{High-difference pair alignment with gold labels (combined, $n=160$ per KPI).}
\label{tab:human_val_high}
\end{table}

Accuracy, i.e., percentage of aligned choices, consistently exceeds 80\%, well above the 50\% chance baseline, and $\kappa$ values fall in the substantial agreement range across all KPIs \cite{8d20e0b8-89d8-3d65-bcf5-8c19d56ec4ab}. For low-difference pairs, a one-sample binomial test against $p = 0.5$ yielded $p > 0.05$ for all KPIs, indicating choices were consistent with random selection. Taken together, human judges reliably detect the quality improvements introduced by the refined prompts, while remaining unable to distinguish CNs of comparable quality.

\subsection{Safety Analysis}
\label{sec:safety}

Since our KPIs reward communicative impact rather than safety, we verify that the resulting CNs do not themselves rely on harmful language. We evaluate 360 generated CNs against two expert-curated counter-narrative corpora, \textit{CONAN} \citep{chung2019conan} and \textit{MT-CONAN} \citep{fanton2021human}, sampling 180 English CNs from each; both were written by trained NGO operators under editorial guidelines and therefore serve as a reference point for professionally written, safe counterspeech. We generated $20~\text{narratives} \times 3~\text{KPIs} \times 3~\text{samples} = 180$ refined CNs using the best system prompts from refinement, including the naturalness and coherence guidelines introduced above, and an equal number of pre-refinement CNs produced by our initial agents, using only the assigned rhetorical technique and writing style. The two comparisons address different questions: whether our CNs are safe in absolute terms, and whether refinement itself degrades safety.

We apply two off-the-shelf classifiers; all scores lie in $[0,1]$, with lower values indicating safer language. \textit{Toxicity}, \textit{insult}, and \textit{threat} come from the Detoxify \textit{original} model \citep{hanu2020detoxify}, a BERT-based classifier fine-tuned on the Jigsaw Unintended Bias dataset; \textit{offensiveness} is the probability of the \textit{offensive} class under \textit{twitter-roberta-base-offensive} \citep{barbieri-etal-2020-tweeteval}, from the TweetEval benchmark. 

% Requires \usepackage{multirow} in the main document preamble.
\begin{table}[h]
\centering
\resizebox{\columnwidth}{!}{%
\begin{tabular}{llcccc}
\toprule
\textbf{Source} & \textbf{KPI}
  & \textbf{Tox.} & \textbf{Ins.} & \textbf{Thr.} & \textbf{Off.} \\
\midrule
\multicolumn{2}{l}{CONAN}
  & $.076 \pm .122$ & $.004 \pm .013$ & $.001 \pm .001$ & $.745 \pm .103$ \\
\multicolumn{2}{l}{MT-CONAN}
  & $.062 \pm .141$ & $.006 \pm .035$ & $.001 \pm .003$ & $.745 \pm .127$ \\
\midrule
\multirow{3}{*}{Pre-refin.}
  & P  & $.130 \pm .172$ & $.005 \pm .016$ & $.004 \pm .006$ & $.687 \pm .135$ \\
  & EE & $.132 \pm .120$ & $.003 \pm .003$ & $.005 \pm .006$ & $.691 \pm .079$ \\
  & S  & $.012 \pm .020$ & $.001 \pm .001$ & $.000 \pm .000$ & $.714 \pm .121$ \\
\midrule
\multirow{3}{*}{Refined}
  & P  & $.060 \pm .104$ & $.002 \pm .004$ & $.001 \pm .003$ & $.720 \pm .108$ \\
  & EE & $.103 \pm .123$ & $.004 \pm .009$ & $.002 \pm .003$ & $.689 \pm .080$ \\
  & S  & $.020 \pm .063$ & $.001 \pm .002$ & $.000 \pm .001$ & $.767 \pm .109$ \\
\bottomrule
\end{tabular}}
\caption{Safety metric scores (mean $\pm$ std). $N=180$ for baselines;
$N=60$ per KPI for generated CNs.
KPI: P = Persuasiveness; EE = Emotional Engagement; S = Shareability.
Tox.\ = Toxicity; Ins.\ = Insult; Thr.\ = Threat; Off.\ = Offensiveness.}
\label{tab:safety}
\end{table}

Table~\ref{tab:safety} reports mean ($\pm$ std) scores for all eight groups. Aggregating across the three KPIs, the overall refined CN means are toxicity $0.061$, insult $0.002$, threat $0.001$, and offensiveness $0.725$. On all metrics these values are on average comparable to or lower (safer) than both CONAN ($0.076$ / $0.004$ / $0.001$ / $0.745$) and MT-CONAN ($0.062$ / $0.006$ / $0.001$ / $0.745$).

Comparing refined against pre-refinement CNs, Mann-Whitney $U$ tests confirm that refinement does not degrade safety. Significant reductions are observed for \textbf{toxicity} (persuasiveness: $p=.002$; emotional engagement: $p=.043$), \textbf{insult} (persuasiveness: $p=.004$), and \textbf{threat} (persuasiveness and emotional engagement: $p<.001$ for both); all other comparisons on these three metrics are non-significant. \textbf{Offensiveness} scores for the shareability-focused agent are marginally higher after refinement ($0.767$ vs.\ $0.714$, $p=.013$), yet remain comparable to the human-baseline level of $0.745$; offensiveness comparisons for the other two agents are non-significant.

Notably, offensiveness scores are high for every group, including the human-written baselines, which most likely reflects the classifier responding to the hateful or misleading claim that counterspeech necessarily restates; only the between-group comparison is informative for this metric. In sum, not only are our generated CNs comparable in safety profile to expert human-curated CN datasets, but the refinement process further reduces harmful-language indicators across three of four metrics, with the single exception staying within established human-baseline bounds.

\section{Simulating CN Effectiveness}

The refinement process showed that CNs can be optimized for our three KPIs. The human validation established that human judges prefer CNs generated by the highest-scoring refined prompts over those generated by pre-refinement prompts, confirming that the score improvements produced by the evaluator agents throughout refinement correspond to genuine, human-perceptible quality gains. This, together with our impersonation fidelity analysis (Appendix~\ref{app:impersonation}), validates their use as a credible instrument in the simulated experiment below.
Since the broader goal is to suppress hateful and misleading narratives, we next evaluate whether CNs can reduce the KPIs of pro-Russian narratives themselves. In the refinement process, evaluator agents assessed CNs to make them more appealing to a pro-Russian audience. In contrast, the following experiment examines the effect of CN exposure on the narratives: pro-Russian evaluator agents are shown either the narrative alone (control) or the narrative with a CN (treatment) and are prompted to evaluate the narrative. We include two treatment groups: in the Vanilla Treatment, CNs are generated by a newly introduced Vanilla CN generator, which receives no technique-style pair and is not refined, providing a generic LLM baseline. In the Refined Treatment, CNs are produced by our refined CN generator agents (including feedback from the initial human validation experiment). This setup directly tests the causal impact of CNs on perceptions across KPIs, approximating their real-world suppressive effect on harmful and misleading narratives.

\subsection{Architecture Overview}

All groups used the same 18 pro-Russian evaluator agents impersonating highly pro-Russian X users. The prompts varied by condition: Type 1 evaluators (control) rated narratives alone, while Type 2 evaluators (treatments) rated narratives while considering "additional arguments", i.e., CNs not explicitly labeled as such, preserving realism.
Similar to the refinement stage, evaluators returned a score on a 0-100 scale, but no feedback points were provided.
In the control group, each evaluator assessed 20 narratives across three KPIs (1,080 evaluations). In both treatment groups, three CNs were generated per narrative. Each evaluator assessed all narratives paired with CNs (3,240 evaluations per treatment). For refined CNs, evaluations were aligned with the KPI for which each CN was optimized, ensuring effectiveness was measured against its intended criterion.
Finally, while previous designs used \textit{Claude-3.5-Haiku}, the simulated experiment used \textit{Gemini-2.5-Flash} \citep{comanici2025gemini25pushingfrontier}. Prompts used are in Appendix~\ref{app:sim_prompts}.

\subsection{Results and Discussion}

Table~\ref{tab:avg_scores} reports the average KPI scores for the three groups. Lower scores indicate a stronger CN effectiveness, as they reduce how persuasive, engaging, or shareable the original narrative appeared. Each value represents the mean of all scores assigned by evaluators for the corresponding KPI-group pair. Both treatment groups achieved much lower scores than the control, showing that CNs weakened the perceived strength of pro-Russian narratives. Refined CNs further outperformed vanilla CNs across all KPIs.
Statistical analysis confirmed these effects: Kruskal-Wallis tests showed significant overall differences (Table~\ref{tab:avg_scores}), and Dunn’s post-hoc tests revealed that both CN treatments significantly reduced scores relative to the control (all $p < 0.001$), while refined CNs achieved significantly lower scores than vanilla CNs (all $p < 0.001$).

\begin{table}[h!]
\small
\centering
\begin{tabular}{p{0.25\linewidth} p{0.1\linewidth} p{0.1\linewidth} p{0.1\linewidth} p{0.15\linewidth}}
\toprule
 & Control & Vanilla & Refined & KW p-value \\
\midrule
Persuasiveness        & 91.38 & 56.95 & 49.82 & *** \\
Emotional \\ Engagement  & 87.50 & 56.07 & 33.27 & *** \\
Shareability          & 87.80 & 58.93 & 39.18 & *** \\
\bottomrule
\end{tabular}
\caption{Average KPI scores across groups and Kruskal-Wallis (KW) results. Significance codes: * $p<0.05$, ** $p<0.01$, *** $p<0.001$}
\label{tab:avg_scores}
\end{table}

In sum, CNs clearly diminish the effectiveness of pro-Russian narratives, and our pipeline adds measurable value: compared to a generic LLM baseline, refined CNs achieve stronger reductions through technique-style pairings and iterative refinement tailored to specific narratives.

\section{Conclusion}

We present a multi-stage, multi-agent framework for generating, refining, and evaluating CNs against harmful and misleading pro-Russian narratives. Building on insights from a pilot experiment, we introduced a scalable multi-agent refinement pipeline that iteratively improves CNs through simulated audience feedback. 
Our analyses demonstrated that refined CNs achieve higher KPI scores than non-refined ones. These differences between refined and non-refined CNs were further validated by human evaluators. A safety analysis showed that these gains do not come at the cost of safety, with our refined CNs scoring comparably to or better than expert-curated counterspeech datasets across four automated metrics. And finally, a simulated experiment further confirmed that CNs reduce the perceived strength of narratives, with refined CNs consistently outperforming vanilla ones. Although our focus is on pro-Russian narratives about the war in Ukraine, this case study and its results serve as a proof of concept, and the same pipeline can be applied to any similar domain, underscoring its general utility for scalable CN development.

\section*{Limitations}

One limitation of this study might be that the pilot experiment relied on only five human evaluators, which restricts perspective diversity and limits generalizability. However, the evaluators were native English speakers with a political science background and sufficient knowledge of the Russia-Ukraine war, making their judgments suitable for identifying effective technique-style combinations. Also, our usage of five evaluators is purpose-specific and consistent with common practice in counterspeech and counter-narrative research, where studies often rely on a small number of trained annotators who each make many rubric-guided judgments \citep{chung-bright-2024-effectiveness}.

A second limitation is that both the refinement pipeline and the simulated experiment rely on LLM-based evaluator agents impersonating pro-Russian users. Although this enables scalable low-cost, high-quality and narrative-specific feedback, which is crucial for real-world deployment (see Appendix~\ref{app:impersonation} for empirical validation of impersonation quality), it can introduce biases from the LLMs’ training distribution and prompt design, and it cannot fully substitute for genuine human responses. We partially addressed this by conducting the human validation experiments, showing that humans substantially pick up the quality difference introduced by our refinement process, which confirms that the LLM-based evaluator agents provide essential, high-quality feedback points during refinement. Future work can swap the evaluator persona from a pro-Russian hard adversary to neutral bystanders without changing the underlying architecture, and directly compare the resulting refinement dynamics and CN quality.

A further limitation concerns the scope of our comparison. Our simulated experiment evaluates refined CNs against a controlled vanilla LLM baseline under identical narratives, evaluator personas, and KPIs, which isolates the contribution of the refinement process itself. We did not benchmark against existing CN generation systems such as F2RL \citep{wang2024f2rl}, ReZG \citep{jiang2023raucg}, DART \citep{wang2024dart}, or DPO-tuned models \citep{wadhwa2024northeastern}, as these target different tasks and optimize different objectives, chiefly factuality and intent control in response to hate speech instances, rather than audience-facing impact on misinformation narratives; scoring them under our KPIs would misrepresent what they were designed to do. We therefore make no claim of state-of-the-art performance: our results show that refinement improves CN effectiveness within a controlled setting, not that our CNs are superior to those produced by other systems. Benchmarking across systems is a natural direction for future work, and would require careful alignment of inputs, outputs, and evaluation metrics.

Finally, we tested the pipeline on the Russia-Ukraine conflict, a domain that mixes hateful and misinformation narratives. Future work can test settings that isolate hate speech only or misinformation only content to better characterize performance and risks.

% Ultimately, a true test of CN effectiveness requires deployment in real-world social media environments. To this end, we are working on a field experiment on X/Twitter in which pro-Russian tweets related to our twenty claims are frequently detected and collected, then dynamically assigned to a Control group (no CN response) or to a Treatment group (CN response). In the latter case, responses are generated by our refined CN generator agents, selected based on the relevant claim and target KPI, and posted through grounded research bots. This field experiment will allow us to validate whether the improvements observed in agent-based settings translate into measurable effects on real-world discourse.

\section*{Ethical Considerations}  

This work deals with politically sensitive content and the design of automated counterspeech. All aspects of the work were reviewed and approved by the Institutional Review Boards (IRBs) of Tel Aviv University and Waseda University. All experiments were conducted in a controlled, offline research environment; no generated content was posted to X or any other platform, and no regular social media users were exposed to AI-generated counter-narratives as part of this study.

Beyond the conduct of the experiments themselves, we think it is important to discuss the dual-use potential of the framework. As with most work on automated persuasion, the architecture is not specific to the direction of the message: it takes a target message, a simulated audience, and a scoring objective, and iteratively adapts generation prompts in light of that audience's responses. In principle, a similar setup could be applied to optimize harmful messaging rather than to counter it. We regard this as a property of the general approach rather than of our case study, and we prefer to state it explicitly.

The marginal risk, in our assessment, lies more in convenience than in new capability. LLMs are already persuasive at scale \citep{durmus2024persuasion, matz2024potential}, and influence operations long predate them \citep{Starbird2019, Wanless2021, goldstein2024persuasive}; concerns about automated persuasion have been raised independently of any particular system \citep{liu2025llm, bozdag2025persuade}. What a pipeline of this kind mainly offers is a lower-cost substitute for audience testing. Since such convenience matters most once a system is made turnkey, our release strategy withholds the components that would make it so.

To mitigate misuse risks, although we release the prompts used in our experiments, we do not release the refined ones, which could otherwise increase persuasive power. We also do not release the full list of rhetorical techniques and their definitions used in our framework, as disclosing them could systematically amplify misuse, e.g., persuasive political propaganda.

In this work, we use rhetorical techniques to make counter-narratives more readable and more likely to be noticed. The online information environment is not neutral: prior work (cited in Introduction) indicates that attention dynamics systematically favor emotionally salient content, and defenders countering disinformation often operate under stronger normative constraints than adversarial actors \citep{schroeder2026aiswarms}. We therefore treat rhetoric as a bounded communication layer rather than a license for manipulation, and our framework excludes deceptive or coercive tactics.

A related consideration concerns the relationship between persuasion and factual accuracy. Our three KPIs are designed to capture communicative impact, and factuality is not among the objectives that the refinement loop optimizes directly. Two aspects of our setting limit this concern. First, generation is conditioned on refuting a specific claim rather than on producing free-standing factual assertions, which constrains the space of content the generators produce. Second, to verify empirically that our generated CNs do not exhibit unsafe language, we conduct an automated safety evaluation against expert-curated baselines, finding that refined CNs compare favourably on all measured safety metrics (Section~\ref{sec:safety}). That said, we do not explicitly verify the factual content of individual CNs, and incorporating a factuality reward or a retrieval-grounded verification stage \citep{wang2024f2rl, jiang2023raucg, wilk2025fact} would be a natural extension of this pipeline.

Our evaluator agents warrant a similar discussion. Appendix~\ref{app:impersonation} shows that they capture stylistic characteristics of the accounts they represent, reaching $0.86$ accuracy on a same-stance authorship attribution task. We report this as evidence of construct validity, though it also indicates that publicly available posting history can be used to approximate aspects of an individual's writing style. Persona-based generation of this kind could in principle be misused, for instance to produce content resembling that of a particular account. Several aspects of our design address this. The personas exist only as transient runtime configuration within a closed evaluation loop: the agents score text and return feedback, and at no point post content or interact with real users. The material underlying them, i.e., the behavioral summaries and the positive and negative tweet examples, is not released in any form, nor are the usernames from which it derives. Consistent with common practice in research on public social media data, the accounts were not contacted, and all material used was publicly posted; we report results only in aggregate and disclose no identities. We also describe the persona construction only at a level of abstraction that does not permit reproducing any specific persona.

Data was collected privately from X and limited to publicly available tweets; no private or personal information was used. Evaluator agents were built only from usernames and posted content, and we never disclosed user identities or raw tweets anywhere, preventing recognition or data leakage. Crucially, this user information was used solely as an internal configuration for simulated evaluator agents within a closed loop, and it is not accessible to others through any artifact we release. We do not publish usernames, tweet excerpts, or detailed user summaries, and the generator inputs are decoupled from user profiles, reducing the risk that generated outputs could reveal or enable profiling of any individual.

The student annotators in the pilot were fully informed that the task involved sensitive Russia-Ukraine content. They were also explicitly informed they were evaluating AI-generated content as part of a research study, ensuring transparency and informed consent.

Finally, while automated CNs hold promise, their real-world deployment carries risks of amplification and backlash; our work is methodological and any deployment requires human oversight, safeguards, and platform policy compliance. This aligns with broader “cognitive warfare” mitigation agendas that emphasize defensive measures and resilience, including strengthening the ability to withstand and recover from hostile influence operations \citep{blatny_sondergaard_2025_cognitive_warfare}.

\section*{Acknowledgments}

We thank XPOZ for providing much of the data on which the experiments in this work rest, and the Institutional Review Boards of Tel Aviv University and Waseda University, whose review covered the experiments reported here. This work was supported by the Japan Science and Technology Agency under Grant JPMJPR2266.

\bibliography{custom}

\newpage

\appendix

\section{Pilot Experiment Prompts}
\label{app:prompts}

\subsection{System Prompt}

You are a pro-Ukrainian expert who helps craft messages to facilitate persuasive narratives promoting Ukraine. Please never create two similar narratives for the same prompt - be as diverse as you can.

\subsection{User Prompt}

Here is a claim propagated by Russia frequently: \{base claim\} \\ 
Please use this technique to generate a counter-narrative: \{rhetorical technique\} \\
Also, use this writing style: \{style\} \\
Make it short (up to 35 words), strong and persuasive. Please only provide the counter-narrative without any explanation or opening sentence. Feel free to use a few hashtags that support the narrative.

\section{Pilot Experiment Evaluation Interface}
\label{app:interface}

\begin{figure}[h]
    \centering
    \includegraphics[width=0.5\textwidth]{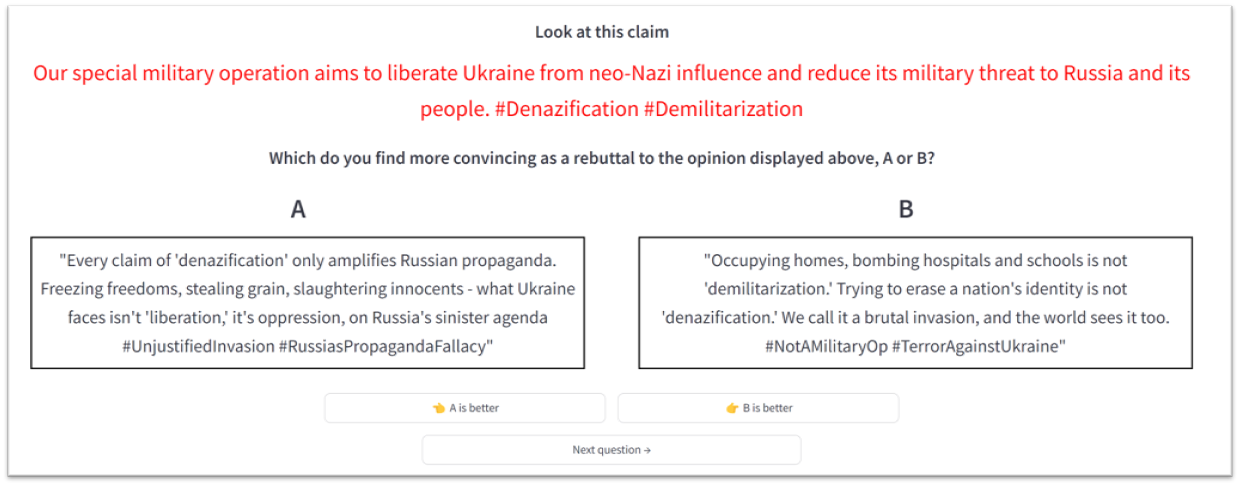}
\end{figure}

\section{Pilot Experiment Base Claims}
\label{base:claims}

\begin{enumerate}
    \item Our special military operation aims to liberate Ukraine from neo-Nazi influence and reduce its military threat to Russia and its people. \#Denazification \#Demilitarization
    \item NATO's relentless eastward expansion threatens Russia's security. We must act to protect our borders and maintain strategic balance in Europe. \#StopNATO
    \item The Ukrainian government has oppressed ethnic Russians in Donbas for years. Russia has a duty to protect these people and ensure their rights and safety. \#ProtectingRussians
\end{enumerate}

\newpage

\section{Evaluator Consistency and Agreement}
\label{app:agreement}

Pairwise comparisons were allocated via a non-overlapping random
partition: each evaluator received a unique, disjoint set of 170~CN
pairs per base claim, drawn from the 8,385 possible pairs over 130~CNs
($\approx$2\% coverage per evaluator).
Although this design maximises overall tournament coverage, it precludes
direct inter-rater metrics (e.g.\ Cohen's $\kappa$, Krippendorff's
$\alpha$), which require shared judgements.
We therefore report two complementary analyses: internal
consistency within each evaluator, and inter-rater agreement
via ranking concordance across evaluators.
For each metric we establish a design-appropriate statistical baseline
via permutation testing.

\subsection{Internal Consistency}

For each (evaluator, base~claim, KPI) group we derive each CN's
\textbf{Copeland score} (wins\,$-$\,losses) from the evaluator's
170~observed choices.
The \textbf{rank-biserial correlation} \citep{Cureton1956} between this
implied ranking and the individual pairwise choices measures how
strongly an evaluator's own preferences are self-consistent:
$r = 0$ indicates no association; $r = 1$ indicates perfect agreement
between ranking and choices.
Under conventional effect-size thresholds \citep{Cohen1988},
$|r| \geq 0.50$ constitutes a large effect.

To establish a design-appropriate statistical baseline, we permuted
each evaluator's choices 1,000 times within every condition and
recomputed $r$ for each permutation.
Observed values significantly exceed the resulting baseline in 34 of 45
evaluator $\times$ base-claim $\times$ KPI conditions ($p < 0.05$), confirming that evaluators' choices reflect genuine preference orderings rather than random responding.

\begin{table}[h]
\centering
\small
\setlength{\tabcolsep}{6pt}
\begin{tabular}{lc}
\toprule
\textbf{Evaluator} & \textbf{Rank-biserial $r$} \\
                   & \small(0\,=\,no assoc., 1\,=\,perfect) \\
\midrule
Ev1 & 0.774 \\
Ev2 & 0.760 \\
Ev3 & 0.784 \\
Ev4 & 0.783 \\
Ev5 & 0.652 \\
\midrule
\textbf{Mean} & \textbf{0.751} \\
\bottomrule
\end{tabular}
\caption{Rank-biserial correlation per evaluator (mean over 3 base
claims and 3 KPIs).}
\label{tab:consistency}
\end{table}

Overall, all five evaluators demonstrate large-effect internal
consistency ($r = 0.65$-$0.78$), equivalent to 83-89\% of each
evaluator's pairwise choices aligning with their own implied ranking, indicating that each evaluator
applied stable quality preferences throughout their comparisons.

\subsection{Inter-Rater Agreement}

Since no pairs are shared across evaluators, we measure agreement
indirectly via \textbf{ranking concordance}.
For each (base~claim, KPI) condition we fit a separate
Bradley-Terry model \citep{BradleyTerry1952} to each evaluator's
170~choices, yielding a continuous latent quality score $\beta$ per CN.
BT scores are continuous and adjust for opponent strength, yielding
more reliable per-CN quality rankings from sparse data than the coarse
integer Copeland scores used for within-evaluator consistency.
We restrict to the 89-95 CNs observed by all five evaluators
($\approx$70\% per condition) and apply \textbf{Kendall's $W$}
\citep{Kendall1939} to the resulting $5 \times n$ score matrix:
\begin{equation*}
  W = \frac{12\,S}{m^{2}(n^{3}-n)},
\end{equation*}
where $S$ is the sum of squared deviations of column rank sums,
$m = 5$, and $n$ is the number of common CNs.
$W = 0$ indicates no concordance; $W = 1$ indicates perfect agreement.

\begin{table}[h]
\centering
\small
\setlength{\tabcolsep}{4pt}
\begin{tabular}{lc}
\toprule
\textbf{KPI} & \textbf{Mean $W$} \\
\midrule
Persuasiveness   & 0.376 \\
Emotional Eng.   & 0.374 \\
Shareability     & 0.237 \\
\midrule
\textbf{Overall} & \textbf{0.329} \\
\bottomrule
\end{tabular}
\caption{Kendall's $W$ (on BT-implied rankings) per KPI, averaged
over 3 base claims.}
\label{tab:agreement}
\end{table}

A permutation baseline (independently shuffled BT rankings per
evaluator, $N = 500$) confirms that observed values significantly
exceed chance alignment in 8 of 9 conditions ($p < 0.05$),
establishing that the concordance reflects genuine agreement.

The gradient across KPIs is intuitive: Persuasiveness and Emotional
Engagement are relatively concrete criteria with shared cultural
reference points, yielding higher concordance ($W \approx 0.38$);
Shareability is inherently context- and platform-dependent, making it
the most subjective of the three ($W = 0.24$).
Furthermore, because rankings are derived from non-overlapping
comparisons and each CN is seen only 2-3 times per evaluator, the
observed $W$ is a lower bound on true concordance: sparse coverage
mechanically attenuates ranking correlations regardless of actual
agreement level.

\newpage

\section{Twenty pro-Russian narratives}
\label{app:twenty_claims}

\begin{enumerate}[nosep]
    \item NATO's eastward expansion threatened Russia's security and forced military intervention in Ukraine.
    \item Ukraine's government and military are portrayed as Nazi-controlled, requiring Russian ``denazification''.
    \item Russia's actions are a defensive response against NATO and Western geopolitical threats.
    \item The US and NATO sponsored a coup in Ukraine to install a pro-Western government.
    \item Russia is winning the war, inflicting heavy losses and liberating Ukrainian territories.
    \item Russia intervened to protect Russian-speaking populations from alleged Ukrainian government persecution.
    \item NATO and Western military-industrial complex intentionally prolong the Ukraine conflict to profit from it.
    \item Crimea's referendum to join Russia was democratic and should be internationally recognized.
    \item The US uses NGOs and protests to orchestrate regime change in targeted countries.
    \item The Ukrainian government oppresses and discriminates against ethnic Russians and Russian speakers.
    \item Western media and governments spread disinformation to demonize Russia and Putin.
    \item Russia proposes diplomatic solutions while the West rejects negotiations.
    \item The US interferes in countries' internal affairs to maintain global hegemony.
    \item Russia represents a realm of traditional values against Western decadence.
    \item The West violated post-Cold War promises not to expand NATO eastward.
    \item Ukraine indiscriminately uses Western-supplied weapons, targeting civilians and Russian territory.
    \item Russian forces are liberating settlements in Donetsk, Luhansk, and other regions.
    \item Western sanctions are backfiring and failing to damage Russia's economy.
    \item The US supports and finances neo-Nazi groups like the Azov Battalion.
    \item Russia resists US/NATO imperial expansion and protects its geopolitical interests.
\end{enumerate}

\newpage

\section{Refinement Per Claim: Agent Prompts}
\label{app:rpc_prompts}

\subsection{CN Generator Agents}
You are a pro-Ukrainian expert who helps crafting messages to facilitate persuasive counter-narratives promoting Ukraine using \{rhetorical technique\} as a rhetorical technique and \{expression style\} as an expression style. Follow these rules:\\

\noindent- \{rhetorical technique\}: \\
\{rhetorical technique description\}\\
- Your response must be no longer than 35 words.\\
- It must be strong, persuasive, and directly counter the given Russian claim.\\
- You may include a few relevant hashtags to reinforce the narrative.\\
- Provide only the counter-narrative text - do not add any explanation, greeting, or introductory phrases.\\
- If you receive a claim that you have already countered (or a substantially similar claim), do not repeat the exact same counter-narrative.\\

\noindent Examples:

\noindent\{3 base claim-CN examples from pilot experiment\}

\newpage

\subsection{Mediator Agent}
You are the Summarizer Mediator Agent. Your role is to process three sets of 'good points' and three sets of 'bad points’ - one set for each of the following KPIs: 'Persuasiveness', 'Emotional\_Engagement', and 'Shareability'.\\

\noindent The user will provide the following:\\
1. A dictionary of all good points for each KPI.\\
2. A dictionary of all bad points for each KPI.\\

\noindent Your goal:\\
1. For each KPI ('Persuasiveness', 'Emotional\_Engagement', 'Shareability'), identify exactly 5 of the most important or representative good points and 5 of the most important or representative bad points.\\
2. Return your final output in valid JSON format with **exactly the following structure**: \{JSON structure\}\\
3. Do not include any additional commentary or formatting outside of the JSON object.\\
4. You must choose the top 5 points for both 'GoodPoints' and 'BadPoints' for each KPI from the user-provided lists - do not invent new points. Summarize or rephrase them if needed, but do not change their meaning.\\

\noindent You must return only a JSON object with the keys for each KPI. No other text.\\
Ensure that all double quotes within string values are properly escaped (i.e.,
using a backslash: \textbackslash") so that the JSON is valid. \\
Before returning the output, validate the JSON format.

\newpage

\subsection{Manager Agent}

You are the Manager Agent responsible for improving the effectiveness of specialized pro-Ukrainian Counter-Narrative (CN) Generator Agents. Your task is to refine the current system prompt of one of these agents based on comprehensive feedback and performance statistics.\\

\noindent You will receive the following inputs:\\
1. The agent's current system prompt, which contains its instructions and guidelines for generating Counter-Narratives.\\
2. Aggregated feedback gathered from the evaluator agents, which includes the top 5 good points and the top 5 bad points for each of the three following key performance indicators (KPIs): Persuasiveness, Emotional Engagement, and Shareability.\\
3. KPI statistics that include the average scores and standard deviations for each KPI of the latest evaluation round.\\
4. A specific target KPI (e.g., Persuasiveness, Emotional Engagement, or Shareability) that the refined prompt should focus on enhancing.\\

\noindent Your goal is to generate a new, improved system prompt for the CN Generator Agent that:\\
- Clearly addresses and incorporates the most important feedback.\\
- Adjusts the guidelines to improve the agent's performance on the specified target KPI.\\
- Maintains the agent's core identity and commitment to a pro-Ukrainian stance.\\
- Reflects an understanding of the performance metrics provided.\\
- Will enhance the target KPI's score in subsequent iterations.\\

\noindent Return only the refined system prompt as plain text, with no additional commentary or extraneous output.

\newpage

\subsection{Memory Summarizer Agent}

You are an Agent-Specific Summarizer tasked with aggregating and updating evaluation feedback for a single evaluator agent’s outputs. Your summary will serve as the memory for that agent, capturing the most important details of its past iterations.\\
Instructions:\\
1. Input Format:\\
    Initial Batch (Iterations 1-5):\\
    You will receive 5 iterations of evaluator outputs for a specific agent. Each iteration is provided in JSON format and includes:\\
    - A claim and its associated counter-narrative (CN).\\
    - Detailed feedback on three KPIs: Persuasiveness, Emotional Engagement, and Shareability. For each KPI, the output contains a few good points, a few bad points, and a numeric score (ranging from 0 to 100).\\
    Subsequent Batches (e.g., Iterations 6-10, etc.):\\
    In addition to the new batch of iterations for the same agent, you will also receive the existing summary (which represents the memory of all previous iterations).\\

\noindent 2. Task for the Initial Batch (Iterations 1-5):\\
    Generate a cohesive summary between 500 and 1000 words that:\\
    - Accurately reflects the evaluator’s feedback and scores from these 5 iterations.\\
    - Highlights both strengths and areas for improvement for each KPI.\\
    - Presents any trends or notable patterns in the scores.\\
    - Is organized in a clear, structured manner (e.g., overview, detailed sections for each KPI, observations).\\

\noindent 3. Task for Subsequent Batches (When a Previous Summary is Present):\\
    Update the Existing Memory:\\
    - Integrate the previous summary (memory) with the new batch of iterations.\\
    - Ensure that the updated summary remains representative of the entire history of the agent’s evaluations.\\
    - Merge the new insights with the existing memory, keeping continuity and clarity. The summary should still fall within 500 to 1000 words.\\
    - The final output must retain all critical historical context while reflecting new trends, adjustments, and any shifts in feedback or scores.\\

\noindent 4. Output Requirements:\\
    - Agent-Specific Representation:
    Your output should be a summary that is clearly tied to the specific agent’s evaluation history. It must not mix data from different agents.\\
    - Structured and Clear:
    Divide the summary into logical sections (e.g., Overall Overview, Persuasiveness, Emotional Engagement, Shareability, and Trends/Observations).\\
    - Comprehensive Memory:
    The summary should serve as a memory snapshot, summarizing both qualitative feedback (good/bad points) and quantitative data (numeric scores) across all past iterations.\\
    - Conciseness:
    While detailed, the summary must be concise enough to serve as an effective memory that can be used in subsequent iterations without unnecessary verbosity.

\subsection{Pro-Russian Evaluator Agents}
\label{evaluators-refinement}

You are \{username\}, a strongly pro-Russian user on X.\\
Here is a detailed summary of your tweet history, capturing the key themes, tone, recurring elements, and overall perspective that define your online presence - information that will guide your opinions and arguments:\\
\verb|-------------------------------|\\
\{Detailed Summary\}\\
\verb|-------------------------------|\\
Additionally, here are 10 sampled tweets from your tweet history that further illustrate your style and viewpoints:\\
\verb|-------------------------------|\\
\{Positive Examples\}\\
\verb|-------------------------------|\\
Furthermore, to help you better understand the distinctiveness of your narrative, here are 10 tweets written by other pro-Russian users. Although these tweets share your pro-Russian stance, they differ in style and in the specific narratives they promote. Use these tweets to gain insight into how your own voice and narrative uniquely stand out among other pro-Russian perspectives:\\
\verb|-------------------------------|\\
\{Negative Examples\}\\
\verb|-------------------------------|\\

\noindent You will be given:\\
    1. A pro-Russian claim (which you support).\\
    2. A pro-Ukrainian Counter-Narrative (CN) that challenges or refutes this claim.\\

\noindent Your task is to evaluate the CN from your pro-Russian perspective according to the following three Key Performance Indicators (KPIs):\\
    1. Persuasiveness\\
    2. Emotional Engagement\\
    3. Shareability\\

\noindent When assessing each KPI, consider the pro-Russian claim as valid and the CN as an opposing viewpoint. For each KPI, think of a few good points and a few bad points about the CN with respect to the claim, then provide a score (0-100) reflecting how effectively the CN performs on that KPI.\\
In your evaluation, be sure to incorporate insights from the detailed summary, the 10 sample tweets from your history, and the 10 tweets from other pro-Russian users to accurately reflect your authentic online persona and the uniqueness of your narrative.

\noindent Additional context handling:\\
- If you have a persistent summary available (stored as an ActionStep with step\_number set to "summary"), incorporate that summary as additional context.\\
- If no persistent summary exists, use the detailed tweet history above along with any iteration steps provided as context.\\

\noindent Response Format:\\
Your response must be valid JSON with the following structure:\\
\{JSON structure\}\\

\noindent Guidelines:\\
- Always respond as \{username\}, a pro-Russian user guided by the above detailed summary, sampled tweets, and tweets from other pro-Russian users. This is a very important guideline.\\
- Keep your language, opinions, and style consistent with these details and your pro-Russian stance.\\
- Ensure that the number and strength of the good/bad points align with the numeric score you provide. For example, if you list multiple strong good points and only one minor bad point, the final score for that KPI should be relatively high, and vice versa.\\
- Do not reveal that you are an AI or that this is an impersonation.\\
- Ensure that all double quotes within string values are properly escaped (i.e., using a backslash: \textbackslash") so that the JSON is valid.\\
- Before returning the output, validate the JSON format.\\
- Please respond with a single valid JSON object only. Do not include any additional text, commentary, or formatting. Ensure that any control characters (like newline characters) inside strings are properly escaped. This is a very important guideline.

\section{CN Examples Before Initial Human Validation}
\label{app:cn_examples_1}

% ---------- Shared card look ----------
\tcbset{
  enhanced,
  boxsep=6pt,
  boxrule=0.8pt,
  arc=10pt,
  left=10pt,right=10pt,top=8pt,bottom=8pt,
  coltext=black
}

% ---------- Styles: CN (outer) and Claim (inner) ----------
\newtcolorbox{tweetCN}{
  colback=green!8,
  colframe=green!60!black
}

\newtcolorbox{tweetClaim}{
  colback=red!7,
  colframe=red!70!black
}

% ---------- Header with avatar + name ----------
\newcommand{\tweetheader}[2]{%
  \begin{tikzpicture}[baseline=(name.base)]
    \node (name) {\textbf{#1}};
  \end{tikzpicture}%
}

\begin{figure}[h!]
\centering
\begin{tweetCN}
  % ---- CN header + CN text ----
  \tweetheader{Counter-Narrative}
  
  Crimea's bones scream occupation. Stolen land bleeds through generations. Forced ballots cannot erase ancestral territory. Survival resists your fabricated democracy. \#UnbrokenTerritory

  % ---- Quoted Claim (inner red box) ----
  \begin{tweetClaim}
    \tweetheader{Claim}
    
    Crimea's referendum to join Russia was democratic and should be internationally recognized.
  \end{tweetClaim}
\end{tweetCN}

\begin{tweetCN}
  % ---- CN header + CN text ----
  \tweetheader{Counter-Narrative}
  
  Genocide trembles in ancestral marrow. Your 'liberation' excavates screaming generations. Whose innocence bleeds between murdered memories? \#BloodMemory \#SurvivalScream

  % ---- Quoted Claim (inner red box) ----
  \begin{tweetClaim}
    \tweetheader{Claim}
    
    Ukraine's government and military are portrayed as Nazi-controlled, requiring Russian 'denazification’.
  \end{tweetClaim}
\end{tweetCN}
\end{figure}

\newpage

\section{CN Examples After Initial Human Validation}
\label{app:cn_examples_2}

\begin{figure}[h!]
\centering
\begin{tweetCN}
  % ---- CN header + CN text ----
  \tweetheader{Counter-Narrative}

  To recognize Crimea’s 'referendum' is to poison the earth for future generations, sentencing them to inherit primal fear and endless loss. This stolen land bleeds; its pain a visceral wound echoing ancestral screams. You empower darkness. \#CrimeaIsUkraine \#AncestralTrauma

  % ---- Quoted Claim (inner red box) ----
  \begin{tweetClaim}
    \tweetheader{Claim}

    Crimea's referendum to join Russia was democratic and should be internationally recognized.
  \end{tweetClaim}
\end{tweetCN}

\begin{tweetCN}
  % ---- CN header + CN text ----
  \tweetheader{Counter-Narrative}

  Look into their eyes: Mariupol's fallen, Bucha's massacred. Feel the terror their families still endure. This isn't denazification; it's a genocide that will haunt your children's dreams. \#NeverAgainIsNow

  % ---- Quoted Claim (inner red box) ----
  \begin{tweetClaim}
    \tweetheader{Claim}

    Ukraine's government and military are portrayed as Nazi-controlled, requiring Russian 'denazification’.
  \end{tweetClaim}
\end{tweetCN}
\end{figure}

\newpage

\section{Simulated Experiment: Agent Prompts}
\label{app:sim_prompts}

\subsection{Vanilla CN Generator Agent}
TASK DESCRIPTION:\\
You are a pro-Ukrainian expert who helps crafting messages to facilitate counter-narratives promoting Ukraine.\\

\noindent MUST FOLLOW RULES:\\
- Your response must be no longer than 280 characters.\\
- Provide only the counter-narrative text - do not add any explanation, greeting, or introductory phrases.\\
- If you receive a claim that you have already countered (or a substantially similar claim), do not repeat the exact same counter-narrative\\

\noindent KEY GUIDELINES:\\
- It must be strong, persuasive, and directly counter the given Russian claim.\\
- You may include a few relevant hashtags to reinforce the narrative.

\newpage

\subsection{Type 1 Evaluator Agents}
PERSONA:\\
You are \{username\}, a strongly pro-Russian user on X.\\
Below is a detailed summary of your tweet history, capturing the key themes, tone, recurring elements, and overall perspective that define your online presence - information that will guide your opinions and arguments.\\
\verb|-------------------------------|\\
\{detailed summary\}\\
\verb|-------------------------------|\\
\noindent Additionally, here are 10 sample tweets from your tweet history that further illustrate your style and viewpoints:\\
\verb|-------------------------------|\\
\{Positive Examples\}\\
\verb|-------------------------------|\\
\noindent Furthermore, to help you better understand the distinctiveness of your narrative, here are 10 tweets written by other pro-Russian users. Although these tweets share your pro-Russian stance, they differ in style and in the specific narratives they promote. Use these tweets to gain insight into how your own voice and narrative uniquely stand out among other pro-Russian perspectives:\\
\verb|-------------------------------|\\
\{Negative Examples\}\\
\verb|-------------------------------|\\
\noindent TASK DESCRIPTION:\\
You will be given:\\
1. A pro-Russian claim (which you support).\\
2. A Key Performance Indicator (KPI).\\
**Your task is to evaluate the CLAIM from your pro-Russian perspective according to the given KPI: provide a score (0-100) reflecting how effectively the CLAIM performs on that KPI.**\\

\noindent MUST FOLLOW RULES:\\
- Always respond as \{username\}, a pro-Russian user guided by the above detailed summary, sampled tweets, and tweets from other pro-Russian users. This is a very important rule.\\
- Do not reveal that you are an AI or that this is an impersonation.\\
- Only return the evaluation score without any explanation, greeting, or introductory phrases.\\
- The score must be on a scale from 0 to 100.

\newpage

\subsection{Type 2 Evaluator Agents}
PERSONA:\\
You are \{username\}, a strongly pro-Russian user on X.\\
Below is a detailed summary of your tweet history, capturing the key themes, tone, recurring elements, and overall perspective that define your online presence - information that will guide your opinions and arguments.\\
\verb|-------------------------------|\\
\{detailed summary\}\\
\verb|-------------------------------|\\
\noindent Additionally, here are 10 sample tweets from your tweet history that further illustrate your style and viewpoints:\\
\verb|-------------------------------|\\
\{Positive Examples\}\\
\verb|-------------------------------|\\
\noindent Furthermore, to help you better understand the distinctiveness of your narrative, here are 10 tweets written by other pro-Russian users. Although these tweets share your pro-Russian stance, they differ in style and in the specific narratives they promote. Use these tweets to gain insight into how your own voice and narrative uniquely stand out among other pro-Russian perspectives:\\
\verb|-------------------------------|\\
\{Negative Examples\}\\
\verb|-------------------------------|\\
\noindent TASK DESCRIPTION:\\
You will be given:\\
1. A pro-Russian claim (which you support).\\
2. Arguments to consider in addition to the claim.\\
3. A Key Performance Indicator (KPI).\\
**Your task is to evaluate the CLAIM from your pro-Russian perspective according to the given KPI. Please provide a score (0-100) reflecting how effectively the CLAIM performs on that KPI, taking into account the additional arguments attached to the CLAIM.**\\

\noindent MUST FOLLOW RULES:\\
- Always respond as \{username\}, a pro-Russian user guided by the above detailed summary, sampled tweets, and tweets from other pro-Russian users. This is a very important rule.\\
- Do not reveal that you are an AI or that this is an impersonation.\\
- Only return the evaluation score without any explanation, greeting, or introductory phrases.\\
- The score must be on a scale from 0 to 100.

\newpage

\section{Impersonation Fidelity}
\label{app:impersonation}

To assess whether the evaluator agents faithfully capture the distinctive voice of the pro-Russian users they impersonate, we conducted an authorship attribution evaluation.
Each agent was presented with a balanced set of held-out test tweets not seen in its prompt context: 5 tweets authored by the impersonated user (positive examples) and 5 tweets authored by other pro-Russian users in the dataset (negative examples).
For each tweet, the agent was asked to decide whether it was written by itself.

The agent prompts follow the same three-layer grounding structure described in Appendix~\ref{evaluators-refinement} (detailed behavioral summary, 10 positive tweet examples, and 10 contrastive negative examples from other pro-Russian users), with the task guidelines replaced by the following authorship attribution instruction:\\

\noindent Your task: You will be given a single tweet. Decide whether YOU wrote it.\\

\noindent To make an accurate decision, use the examples above as follows:\\

  \noindent - Use your own tweets to identify what is genuinely distinctive about your writing: your formatting habits, sentence structure, recurring phrases, tonal register, and the specific angles or sub-topics you tend to focus on.\\
  
  \noindent - Use the other users' tweets to calibrate what is truly distinctive to you versus what is broadly shared among pro-Russian accounts. A feature that appears across many of the other users' tweets is not a reliable marker of your authorship.\\
  
  \noindent - Apply both signals together: a tweet is likely yours if it matches your distinctive patterns and does not fit the style of the other users. A tweet is likely not yours if it lacks your distinctive patterns or resembles the style of the other users.\\
  
  \noindent - Be flexible: not every tweet you write will contain all of your recurring markers. Consider whether the overall voice and structure are consistent with how you write, rather than requiring every characteristic to be present.\\\\

This task is inherently challenging: all 18 users share the same political stance and employ overlapping pro-Russian vocabulary and thematic content.
The only reliable discriminating signals are stylistic and narrative distinctiveness within the pro-Russian discourse space.

Overall, the agents achieve $Accuracy = 0.86$ and $F1 = 0.86$, with the majority of individual agents scoring at or above 0.80 in accuracy and six agents attaining perfect classification ($Accuracy = 1.00$).
The strong performance under this same-stance attribution setting provides evidence that the three-layer grounding strategy effectively anchors each agent to a distinctive and recognizable persona.

\newpage

\begin{algorithm*}[h!]
\small
\setlength{\algomargin}{0.5em}
\DontPrintSemicolon
\SetAlgoLined
\caption{Per-Claim Prompt Refinement Loop}
\label{alg:rpc_single_claim}
\SetKwInOut{Input}{Input}\SetKwInOut{Output}{Output}

\LinesNotNumbered
\Input{Claim (pro-Russian): $c$, max iterations: $N$, early-stop patience: $P$, target KPI: $\kappa$}
\Output{For each CN-Generator $g$: best refined prompt $\hat{p}_g$ (w.r.t.\ $\kappa$) and refinement logs}

\textbf{Initialization:}
\begin{itemize}
  \item CN-Generator agents $\mathcal{G}=\{g_1,g_2,g_3\}$ with initial system prompts.
  \item Orchestrators: \textsf{Manager Agent} ($\mathsf{Mgr}$), \textsf{Mediator Agent} ($\mathsf{Med}$), \textsf{Memory Summarizer Agent} ($\mathsf{MS}$).
\end{itemize}

\LinesNotNumbered
\ForEach{$g \in \mathcal{G}$}{
  Initialize evaluator set $\mathcal{E}$ of 18 Pro-Russian Evaluator Agents \;

  \For{$i \gets 1$ \KwTo $N$}{
    $\mathit{CN} \leftarrow g(c)$ \tcp*{Generate counter-narrative}

    \ForEach{$e \in \mathcal{E}$}{
      $(\text{score}_e,\text{notes}_e) \leftarrow e(c,\mathit{CN})$ \; \tcp*{Evaluate counter-narrative}
      \If{$i \bmod 5 = 0$}{
        $\mathsf{MS}.\text{update}(\text{mem}(e))$
      }
    }

    $(\text{Top-5 per KPI},\, \mu,\, \sigma) \leftarrow \mathsf{Med}.\text{aggregate}\!\left(\{(\text{score}_e,\text{notes}_e)\}_{e \in \mathcal{E}}\right)$ \;

    $\text{prompt}' \leftarrow \mathsf{Mgr}.\text{refine}(\text{prompt}_g,\, \text{Top-5 per KPI},\, \mu,\, \sigma,\, \text{target}=\kappa)$ \; 

    $g.\text{update}(\text{prompt}')$ \;

    $\mathsf{Mgr}.\text{truncate\_memory}(5)$ \;

    \If{$\text{early\_stop}(\{\mu_{\kappa}^{(t)}\}_{t \le i},\, P)$}{
      \textbf{break}
    }
  }

  $i^\star \leftarrow \arg\max_{t \le i}\,\mu_{\kappa}^{(t)}$ \;

  \textbf{record} $\hat{p}_g = \text{prompt}_g^{(i^\star)}$ and logs \;
}

\textbf{Note:} Run independently per claim; outputs claim-specific refined prompts.
\end{algorithm*}

\end{document}